\documentclass[letterpaper, 10 pt, conference]{ieeeconf}  % Comment this line out if you need a4paper

\IEEEoverridecommandlockouts                              % This command is only needed if 
\usepackage{amsmath} % assumes amsmath package installed
\usepackage{graphicx}
\usepackage{subfig}
\usepackage{ulem}
\usepackage{changepage}
\usepackage[mathletters]{ucs}
\usepackage[utf8x]{inputenc}
\usepackage{bm}
\usepackage{xcolor}
\usepackage{cite}
\usepackage{float}
\usepackage{color}
\usepackage{balance}

\title{\LARGE \bf
Conditioned Human Trajectory Prediction using Iterative Attention Blocks
}
\author{Aleksey Postnikov$^{1,2}$ \and Aleksander Gamayunov$^{1,2}$ \and Gonzalo Ferrer$^{2}$%
        \thanks{\textsuperscript{1} The authors are with the Sberbank Robotics Laboratory, Moscow, Russia.
                    {\tt\small \{postnikov.a.l,gamayunov.a.r\}@sberbank.ru}.
               }% <-this % stops a space
        \thanks{\textsuperscript{2}Skolkovo Institute of Science and Technology, Moscow, Russia.
                    {\tt\small g.ferrer@skoltech.ru}.
           }    
    }

\begin{document}

\maketitle
\thispagestyle{empty}
\pagestyle{empty}

%%%%%%%%%%%%%%%%%%%%%%%%%%%%%%%%%%%%%%%%%%%%%%%%%%%%%%%%%%%%%%%%%%%%%%%%%%%%%%%%

\begin{abstract}
    % \textbf{}draft version}
    % Gonzalo: we dont really use robots here... I am going to remove the next two sentences
    %Mobile robots are becoming part of everyday life, they deliver parcels, clean floors and communicate with people at entertainment events. 
    %In these conditions, the requirements for the task of safe and fast movement are growing. 
    Human motion prediction is key to understand social environments, with direct applications in robotics, surveillance, etc.
    We present a simple yet effective pedestrian trajectory prediction model aimed at pedestrians' positions prediction in urban-like environments conditioned by the environment: map and surround agents. % accurate mean positions of pedestrians.
    Our model is a neural-based architecture that can run several layers of attention blocks and transformers in an iterative sequential fashion, allowing to capture the important features in the environment that improve prediction.
    
    We show that without explicit introduction of social masks, dynamical models, social pooling layers, or complicated graph-like structures, it is possible to produce on par results with SoTA models, which makes our approach easily extendable and configurable, depending on the data available. 
    We report results performing similarly with SoTA models on publicly available and extensible-used datasets with uni-modal prediction metrics ADE and FDE. %, which are the most used source of information for the downstream task: motion planning. % GONZALO: this statement we do not demonstrate.
\end{abstract}

% The abstract should answer theses questions:
%    Why did you do it? 
%    What did you do? 
%    How did you do it? 
%    What did you find out? 
%    And what does that mean? 

\section{Introduction}
% Each paper must be divided into two parts. The first part includes the title, authors' name, abstract, and keywords. The second part is the main body of the paper.

% The introduction should answer these questions
%    What is the problem? 
%    Why is it important? (who cares?)
%    Why is it hard?
%    What do we do?

% 1) What is the problem: prediction network
Autonomous mobile robots have recently started walking around us in shops, exhibitions, controlled closed areas around universities and innovative companies. However, human motion is unpredictable by nature, a mind state determining similarly to decision-making and inner motivation processes.
% GONZALO: I am going to remove this sentence. There were some time ago research investigating this issue, social convetions etc, and we really dont try to solve this disambiguation. The interactions is unclear with the datasets we provide, so let's focus
%Are you familiar with the awkward moments when you cannot pass the person that coming to pass you? The path of a robot among people consists entirely of such moments. 
Still, recent data-driven approaches have made a great breakthrough on predicting human trajectories \cite{lstm, sattn, social-lstm, sophie, trajectron, trajectronplusplus} and allow researchers to focus on the most important aspects that might condition such predictions: environment, surrounding agents, past history, or how to interconnect all these components, a task unreachable for past model-based algorithms \cite{ferrer2017,ferrer2014}.

% 2) Why is it important
% GONZALO: I think this paragraph is not helping us, this is one of the applications, but for this paper reading this in paragraph 2 is missleading
%Solving the safety of the robot's movement near a person by using various emergency stop buttons or sensors would vanish the benefits of using robots since they will often stop and do nothing at a safe distance from people. In order to be able to release a mobile robot from a closed area controlled by engineers, robots must independently move in the flow of people, understanding where the person wants to go, and where the robot can freely and safely pave its way.
% Human Motion Prediction is important part of safe and  robot navigation,
%  Human Motion take us open topic to research understanding they decision.

% 3) Why is it important/hard
Regardless of the limitation on predicting human trajectories, the importance of solving human motion prediction is enormous due to its direct applications. To this end, every year, more methods appear, and they are validated on well accepted benchmarks based on open datasets\cite{eth,ucy,sdd}, competing based on common metrics.

% Most of the current algorithms on prediction focus their efforts on predicting accurately the error of state variables
% To this end, multiple benchmarks have been created and released \cite{eth,ucy,sdd}, providing common grounds to test and evaluate.
% Most modern motion prediction algorithms focus on accurate prediction of agent position errors on these benchmarks. Nonetheless, the precision due to this inherent uncertainty is equally important, and this paper is an effort to research on this direction. Prediction algorithms should address this issue as well: it provides a high degree of interpretability by estimating the associated uncertainty and it might be of some use for consequent algorithms making use of prediction information, e.g., planning.

% 4) what do we do? 
In this paper, we focus on a flexible approach of fusing different data input modalities from observed past pedestrian positions and images from the scene.
The flexibility of modern neural-based approaches for the prediction problem allows us to test and verify many different data, such as the perceived environment by map images, the surrounding agents to the ego-agent, etc., and interconnect them in a principled way to obtain the best performance.

\begin{figure}[!h]
    \centering
    \includegraphics[width=8.5cm]{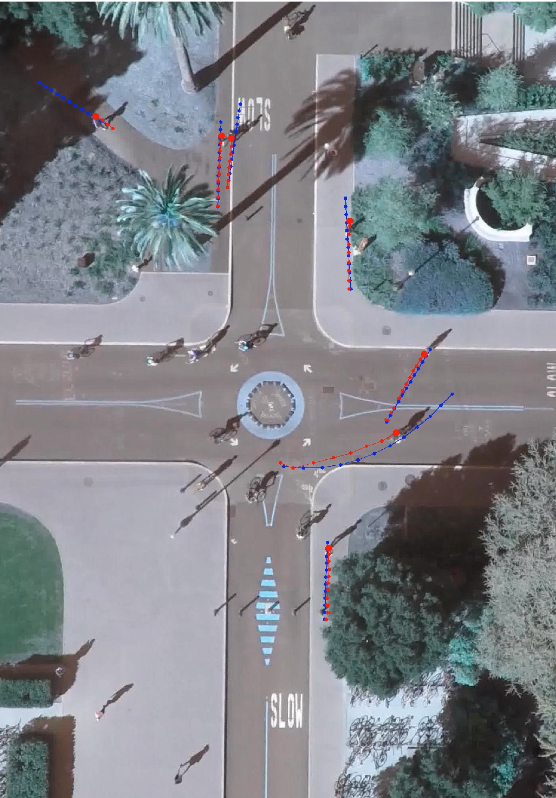}
    \caption{ Visualizations of trajectories predicted by Conditioned  Human  Trajectory  Prediction  using  Iterative  Attention Blocks model (red lines - predictions, blue lines - ground truth trajectories).}
    \label{fig::crop_rot}
\end{figure}

\begin{figure*}[h]
    \centering
    \includegraphics[width=.80\textwidth]{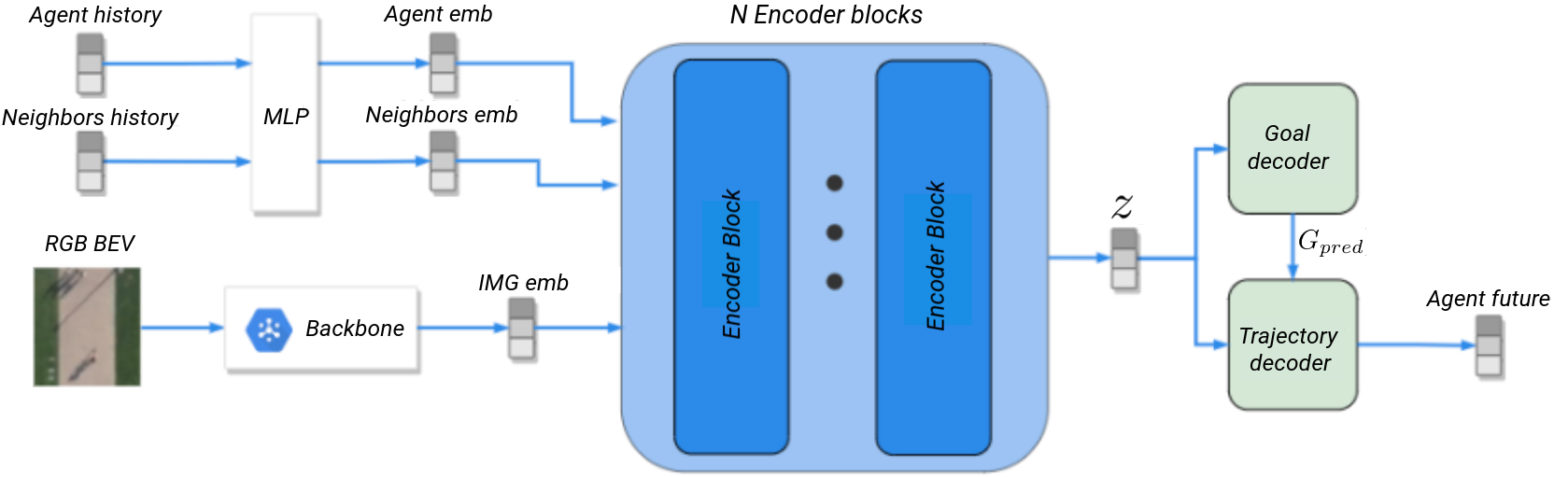}
    \caption{Proposed architecture. Blue blocks is encoder, iteratively extracting information from the input data, green blocks is decoder part divided on two heads: (i) Goal decoder - to predict the motion target and (ii) Trajectory decoder - to predict the trajectory.}
    \label{fig::all}
\end{figure*}

To address this task, we propose (i) a perceiver-like\cite{perceiver}  encoder block, shown on Figure \ref{fig::encoder}, sequentially  mapping data from different modalities with cross-attentions and updating latent vector with  transformers and (ii) goal-conditioned iterative approach for generating trajectory predictions, shown at Figure\ref{fig::all}.

%To address this task, we propose a perceiver-like\cite{perceiver}  architecture, shown on Fig. \ref{fig::all} \gf{explain more, a interleaved network of cross-attention and transformers} and more detailed encoder part on fig.\ref{fig::encoder} \gf{fix this sentence too}.

However, input data preparation is also an essential part of making precise predictions. Therefore, we show preparation the image scene representation before fusing on Figure \ref{fig::crop_rot}, which keeps useful information about agent direction and start position inside.

This paper's main contribution is an approach that interleaves a sequence of cross-attention and transformer blocks to capture better the relations between the high dimensional data embeddings from past histories, neighbors positions and RGB bird-eye view images of the environment. The network design is simplified with respect to other approaches, and the results obtained show an on par performace in the ETH \cite{eth}, UCY \cite{ucy}, and SDD \cite{sdd} datasets.

\section{Related Work}
Motion prediction has become a separate task primarily motivated by the effective navigation of the mobile robot in the social environment. 
Early works were aimed at describing the model of human movement by model-based approach \cite{sfm,hsfm}. 
In some cases of a short prediction horizon or movement on a limited straight sidewalk, the simple linear interpolation or Constant Velocity Model (CVM)\cite{cvm} can show fairly good accuracy. Still, when there are a lot of pedestrians, it is also necessary to take into account their interaction between each other.
This type of interaction can be modeled by Social Force Model (SFM)\cite{sfm} or more related to the human movement model the Headed Social Force Model (HSFM)\cite{hsfm}, which directly take into account such interactions in their model of motion.
However, the accuracy of these models strongly depends on how well the destination point is selected, which is used as input data. 
% In our previous work\cite{sigma-nn} we show how HSFM can predict trajectories using predicted goals by a neural network in a hybrid approach.
\par

In the last years, deep learning approaches have become dominant in the task of Human motion prediction.
Modern deep learning frameworks allow optimizing algorithm computation time by utilizing GPU resources. For example, it has been shown \cite{mpdm, sigma-nn} how empirical models can be packed into layers-like neural network structures that can use optimized gradient descent to calculate backpropagation loss efficiently.
\par
%1 Social LSTM
One of the most popular works that use deep learning approach is Social-LSTM\cite{social-lstm} show a simple but effective way of using Recurrent Neural Networks (RNN)\cite{rnn} variant named Long Short Term Memory (LSTM)\cite{lstm} blocks for learning general human movement and predict their future trajectories.
\par
%1.1 SoPhie
There are another deep learning approach based on Generative Adversarial Network(GAN) named SoPhie\cite{sophie}, which leverages path history information with scene context information, using images. 
% an interpretable framework based on Generative Adversarial Network (GAN)\cite{gan}, which leverages two sources of information, the path history of all the agents in a scene, and the scene context information, using images of the scene. To predict a future path for an agent, both physical and social information must be leveraged.\cite{sophie}
\par
%2 Trajectron++
Another deep learning approach utilizes a graph-structured data model. 
Trajectron \cite{trajectron} and its second version Trajectron++ \cite{trajectronplusplus}  - multi-agent behavior prediction model that accounts for the dynamics of the agents, produces predictions possibly conditioned on potential future robot trajectories which can effectively use heterogeneous data about the surrounding environment.
% copied from icra
\par
PecNet \cite{pecnet} propose estimation of a latent belief distribution modeling the pedestrians’ possible endpoints, which are used to predict trajectories. 
This approach has shown good prediction accuracy, and also it provides additional information about possible trajectories in the outputs, which can be taken into account in trajectory planning step.
\par
% TODO: check this

%  that make few assumptions about their inputs and that
% can handle arbitrary sensor configurations, while enabling
% fusion of information at all levels.
The Perceiver \cite{perceiver} propose novel general perception architectures with latent transformers similar to GPT-2 architecture\cite{gpt2,transformer}, and cross-attention blocks. 
This work is from a different field of image processing, but despite this, we used some of the ideas from this approach, which we adapted for a human motion prediction field.
The main advantage of such an approach is a simplicity and flexibility of architecture while performing on par with the current state of the art\cite{trajectron, trajectronplusplus, sophie}.

% unfinished thought
% We believe that deep learning approaches at an early stage of development now in field of pedestrian prediction and have huge potential in progress. Some ideas may come from more advanced autonomous driving development area
% unfinished thought
\section{Method}
\subsection{Problem Formulation}
%############################################################
% copied from icra 2021   ###################################
% ###########################################################
% added small changes
\par
The position of a generic agent $i \in I$ at time $t$ is represented by $u_i^t = (x,y)$, where $x, y$ are the coordinates of agents in the dataset reference system at the instance of time $t$. The agent's trajectory is defined as $X_i^{1:T} = \{ u_i^1, ... , u_i^T \}$ from timestamp $1$ to $T$.  
% We aim to generate plausible trajectory distributions for a time-varying number of interacting agents.
\par
% $t \in [1,..,T]=[1,..,T_{obs},..,T_{pred}]$
Every trajectory is split into observed and future: given certain number $T_{obs}$ of observed time step positions, and future states or prediction horizon for the next $T_{pred}$ time steps which is denoted as $p( X_i^{T_{obs}+1:T_{pred}}| X_i^{1:T_{obs}})$.
%############################################################
% copied from icra 2021   ###################################
% ###########################################################
\par
% TODO: check common practice links
Following a common practice \cite{cvm,social-lstm,lstm,trajectron} we use $\Delta t=0.4s$ between time steps $t_i$ and $T_{obs}=8; T_{pred}=12$ steps for scene to be predicted, where overall scene time is $(T_{obs}+T_{pred}) \cdot \Delta t=8s$. These constraints are used to unambiguously compare an approach and that can be used to solve a problem with another $\Delta t; T_{obs}; T_{pred}$. 
\par
The task is to predict the next $T_{pred}$ step positions of a pedestrian with minimum differences to ground truth, based on $T_{obs}$ observed position steps and BEV images for all pedestrians in the scene.
\par

\subsection{Data Preparation}
The history of a scene is divided into main agent whose trajectory needs to be predicted $X_m$, where $m \in I$ and neighbors agents $X_{n}$, where $n = \{ i \,|\, i \in I \setminus  m\}$ whose observed trajectories are used as scene context. 
All coordinates $u_i^t$ are normalized relative to the last known position and direction of the main agent $u_m^{T_{obs}}$ with $T_{o \rightarrow m}$ transformation matrix (\ref{eq::mlp_pose_main}, \ref{eq::mlp_pose_neighbors}).
\par
Same multi-layer perceptron (MLP)\cite{mlp} is used to create embedding from agent and neighbors history trajectories:
% (\ref{eq::mlp_pose_main},\ref{eq::mlp_pose_neighbors}).
\begin{equation}
    X_{m}^{emb} = MLP_{pose}(T_{o \rightarrow u_m^{T_{obs}}} \cdot X_{m})
    \label{eq::mlp_pose_main}
\end{equation}
\begin{equation}
    X_{n}^{emb} = MLP_{pose}(T_{o \rightarrow u_m^{T_{obs}}} \cdot X_{n})
    \label{eq::mlp_pose_neighbors}
\end{equation}
Where $T_{o \rightarrow m}$ is transformation matrix from original (dataset) coordinate system to centered to last observed position of agent and rotated along the movement of that agent.
\par
% We use Beard Eye View(BEV) images provided by datasets\cite{sdd,eth,ucy} and enrich them with segmented images by pretrained UNet\cite{unet}. 
% Segmented images are used as an additional source of scene context. 
% We believe \gf{this will be shown below?} that this information might help the model to understand walking surfaces and obstacles along the way in trajectory forecasting process. 
\par
Beard Eye View (BEV) RGB images are rotated along to the last known position and direction of the main agent $u_m^{T_{obs}}$ by applying transformation matrix  $T_{o \rightarrow u_m^{T_{obs}}}$ and cropped after that as shown on Figure \ref{fig::crop_rot} so that last known main agent position always have same image coordinates. 
%$u_m^{T_{obs}} \cdot T_{o \rightarrow m} =(0,0) $
The size of cropping square side $s$ depends on the maximum path length for this type of agent, and it is a hyperparameter of the data preparation module. Agent position centered on image position $[\frac{s}{2},\frac{s}{4}]$.
% In our case we choose $10.0$ meters, and agent  position centered on image position [2.5, 5] meters. but it still our hyperparameter. \textbf{!!rephrase!!}   %TODO: let add some better formulating

\begin{figure}[h]
    \centering
    \includegraphics[width=8.5cm]{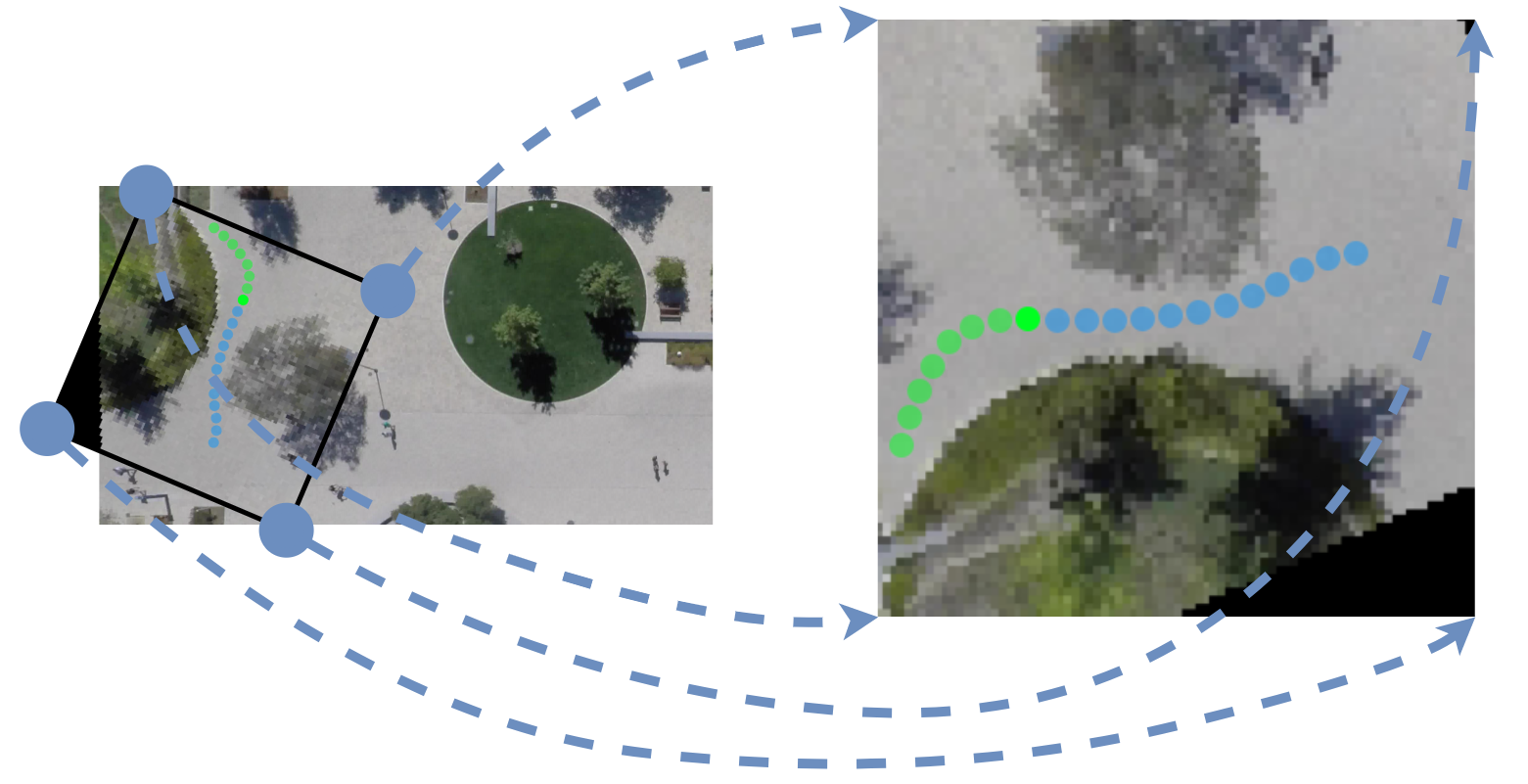}
    \caption{Image rotation and cropping  along the agent position and direction.}
    \label{fig::crop_rot}
\end{figure}

% !first input! same way!manghalam!
We evaluated pretrained Resnet-18\cite{resnet}, ViT-B/16 \cite{vit}, and image patching with linear projections\cite{vit} to create visual features from cropped and rotated BEV RGB  before feeding into encoder. % (\ref{eq::mlp_img}).

% \begin{equation}
%     Img^{emb} = MLP_{img}(Crop(T_{o \rightarrow u_m^{T_{obs}}} \cdot Img ))
%     \label{eq::mlp_img}
% \end{equation}

% \begin{equation}
%     Segm^{emb} = MLP_{segm}(Crop( T_{o \rightarrow u_m^{T_{obs}}} \cdot Segm))
%     \label{eq::mlp_segm}
% \end{equation}

\subsection{Encoder}

\par
We build our architecture from two main components: (i) a cross-attention module that can map latent array and input information of arbitrary shape to a latent array, and (ii) a transformer block that maps a latent array to a latent array.  
Our model applies the (i) cross-attention module and (ii) the transformer in alternation.
This corresponds to projecting the higher-dimensional input information of arbitrary shape through a lower-dimension attention bottleneck before processing it with a deep transformer and then using the resulting representation to query the input again, which allows to effectively fuse different information modalities to the latent array, representing overall scene information.

% The size of the latent array allows us to directly model pixels and to build deeper Transformers, but the severity of the bottleneck may restrict the network’s ability to capture all of the necessary details from the input signal. 

Model architecture is structured with multiple cross-attention layers, which allow the latent array to iteratively extract information from the input data as it is needed. 
% This allows us to tune latent vector size to balance potentially expensive, but informative cross-attends against cheaper, but potentially redundant latent self-attends.
\par
% bad, very bad written
The encoder is composed of a stack of N identical blocks.
Each block has four sub-blocks, namely three cross-attention modules, that map (i) information about agent history, (ii) neighbors history, (iii) RGB environment map, and (iv) latent transformer to update latent vector. 
% The first is ... Generated latent array from $z=N(\mu,\sigma)$ are used as query in 
% \textbf{!!REFORMULATE!!} latent transformer with prepared embedding data (\ref{eq::mlp_pose_main}) as the input of encoder. 
% may be "positional encoding" instead of "time encoding"
The main processing module of used cross attention and latent transformer is attention, which is permutation-invariant by design, which is inappropriate in case of exploiting temporal nature of observed history information. 
To preserve temporal information of history positions, we follow the strategy of Fourier feature position encodings \cite{transformer}.
We concatenate 1D positional(temporal) encodings Eq.(\ref{eq::pe_sin}) \cite{transformer, position_encoding} with agents embedding.  

% RGB and segmentation feature vectors sequentially fed into cross attention parts of encoder as shown in figure \ref{fig::encoder}. 

% ТОДО: check encoding
\begin{equation}
    \begin{aligned}
    PE_{(pos,2i)} = sin(pos\cdot e^{\frac{-4 \cdot i}{d_{model}}})\\
    PE_{(pos,2i+1)} = cos(pos\cdot e^{\frac{-4 \cdot i}{d_{model}}})
    \label{eq::pe_sin}
    \end{aligned}
\end{equation}

% \begin{equation}
%     TE_{(pos,2i+1)} = cos(pos\cdot e^{\frac{-4 \cdot i}{d_{model}}})
%     \label{eq::pe_cos}
% \end{equation}
Where $d_{model}$ - model embedding dimension,
$i\in[0,d_{model})$, specifies model embedding dimension positions,
$pos\in[0,T_{obs})$ - specifies temporal positions
\par
Latent transformer block utilizes the GPT-2\cite{gpt2} architecure, witch is itself the decoder of original Transformer \cite{transformer} architecture.
Cross attention is a multi-head attention layer, Eq. (\ref{eq::multihead}) that decomposed the attention in multiple heads. The independent attention heads are  concatenated and multiplied by a linear layer to match the desired output dimension.
\begin{equation}
    MultiHead(Q,K,V) = Concat(h_1,...,h_i)W^0
    \label{eq::multihead}
\end{equation}
    where \\
    $h_i=Attention(QW_i^Q,KW_i^K,VW_i^V)$, $Q=MLP^Q(k)$, $K=MLP^K(k)$, $V=MLP^V(k)$,
    $W^0=$ output linear block, i - attention layer, k - input 3-dimentional  data.
% Finally we repeat encoder block  N times to increase model depth. 

\begin{figure*}[h]
    \centering
    \includegraphics[width=.80\textwidth]{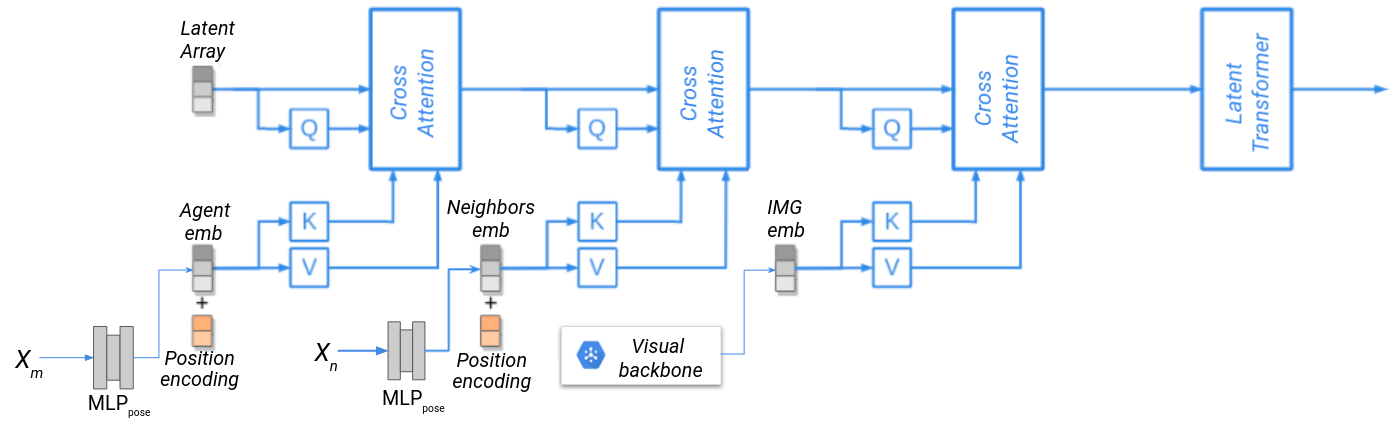}
    \caption{Encoder block. Different input data modalities embedded and sequentially fed into cross attention blocks.}
    \label{fig::encoder}
\end{figure*}

\subsection{Decoder}
\par
We follow the strategy of dividing the trajectory generation problem into two steps: (i) proposing the position of pedestrian goal (goal is pedestrian trajectory point at $T_{pred}$ timestamp), (ii) constructing trajectories conditioned to the proposed goal position.
\subsubsection{Goal Decoder}
\par
We decode final endpoint position \textbf{$u_i^{T_{pred}}$} from encoded latent array, squishing temporal dimension. Multi-Layer Perceptron is used in order to yield our guesses for the final position.
\begin{equation}
    u_i^{T_{pred}} = MLP_{goal}(z)
\end{equation}
where z - latent scene representation vector.

\par
\subsubsection{Trajectory Decoder}
\par
% explain idea of separate train-infer scenarios
% \par
The procedure of Trajectories decoding is different for training and inference scenarios.
During training, Trajectory predictions are obtained by concatenating encoded 3-dimensional latent array with ground truth goals. 
During training, ground truth goals are used because it helps to produce cleaner, less noisy signals for downstream prediction networks while still training the overall module end to end \cite{pecnet}.
% Multi Layer Perceptron is used in order to decode positions from each of temporal dimensions independently(?).
During inference,  predicted goals from Goal Decoder are used instead of ground truth goals. Multi-Layer Perceptron is used to decode trajectory.
\begin{equation}
 u_i^{T_{obs}+1:T_{pred}} = MLP_{traj}(z, G_{gt})   
\end{equation}
\begin{equation}
u_i^{T_{obs}+1:T_{pred}} = MLP_{traj}(z, G_{pred})
\end{equation}
where $G_{gt}$ is ground truth pedestrian position to be predicted at timestamp $T_{pred}$,   $G_{pred}$ is predicted pedestrian position at timestamp $T_{pred}$

\subsection{Loss function}
The proposed architecture has two separate decoders, namely the Goal decoder and Trajectory decoder, which are trained in an end-to-end manner. We use $\mathcal{L}_{FDE}$ to train Goal Decoder and $\mathcal{L}_{ADE}$ to minimize the error of Trajectory decoder:
\begin{equation}
\begin{aligned}
    \mathcal{L}_{ADE} = \frac{ \sum_{t=T_{obs}+1}^{T_{pred}} ||u^t -\mu^t||_2}{T_{pred}}  \\
% \end{equation}
% \begin{equation}
    \mathcal{L}_{FDE} = ||u^{T_{pred}} -\mu^{T_{pred}}||_2 \\
% \end{equation}
% \begin{equation}
    \mathcal{L}_{model} = \mathcal{L}_{ADE} + \lambda \cdot \mathcal{L}_{FDE},
    \label{eq::loss}
\end{aligned}
\end{equation}
where $\lambda$  is a regularizer between two losses.

\section{Evaluation}
\subsection{Datasets}
The model trained and evaluated on ETH-UCY \cite{eth, ucy} with Leave-One-Out Cross-Validation(LOOCV) strategy \cite{loocv} and SDD datasets \cite{sdd}.  
% for accuracy comparison shown in table \ref{table:comparison-eth} and table \ref{table:comparison-sdd}.
% \textbf{!!ADE what is? or pomenyat na loss!!}
% For each case we choose best network weights across $Loss_{model}$ in 40 learning epochs. 
The ETH and UCY\cite{eth,ucy} are publicly available datasets consisted of manual marked pedestrians identifiers and positions on recorded video with 2,5Hz frequency. The videos were recorded from buildings windows and slightly angled, but homography matrices were also attached to datasets for straightening.
\par
The SDD publicly available dataset has identifiers and positions of pedestrians like ETH and UCY, but bigger. It is additionally provides agent class labels such as Pedestrian, Bicyclist, Skateboarder, Cart, Car, Bus, and three flags: ``lost'' - the annotation is outside of the view screen; ``occluded'' The annotation is occluded; ``generated'' The annotation was automatically interpolated.
%SDD dataset recorded from drone, and because of this have field of view bigger than ETH and UCY  with smallest deviation from the horizon.
% \subsection{Train configuration}
% We use $N=12$ encoder blocks, shown on figure \ref{fig::encoder}, with $d_{model}=64$ embedding dimension and drop out rate of 0.1. Our work station configuration allow us to use batch size of 32. 
\subsection{Metrics}
We use Euclidean distance errors: Average Displacement Error (ADE) (\ref{ade}) and Final Displacement Error (FDE) (\ref{fde}) to evaluate the accuracy. Metrics are formulated as: 
% Average Displacement Error is Average L2 distance between the ground truth and our predicted trajectories and Final Displacement Error is the L2 distance between the predicted final destination and the ground truth final destination after the prediction horizon $T_{pred}$.
\begin{equation}
    ADE = \frac{ \sum_{j=1}^N \sum_{t=T_{obs}+1}^{T_{pred}} ||u^t_j -\mu^t_j||_2}{N \cdot T_{pred}}
    \label{ade}    
\end{equation}
\begin{equation}
    FDE = \frac{\sum_{j=1}^N{ ||u^{T_{pred}}_j -\mu^{T_{pred}}_j||_2}}{N}
    \label{fde}
\end{equation}
where N - number of processed pedestrians, $\bm{u}_j^t$ - ground truth position of $\textit{j}^{th}$ pedestrian at timestamp \textit{t}, $\textit{T}_{pred}$ - prediction horizon, $\mu$ - predicted mean position.

\subsection{Compared baselines}
Our method is compared against the following baselines: LSTM\cite{lstm} - LSTM network that process only agent history information by recurrent LSTM layers, S-LSTM\cite{social-lstm} - method with LSTM networks that share the information between the state of agents in a scene through the Social pooling layer, S-ATTN\cite{sattn} - attention-based trajectory prediction model, which uses RNN mixture based approach, modeling both the temporal
and spatial dynamics of trajectories in human crowds, Trajectron++\cite{trajectronplusplus} - graph-structured recurrent model, incorporating agent dynamics and heterogeneous data, PecNet \cite{pecnet} model that infers distant trajectory endpoints to assist in long-range multi-modal trajectory prediction, SoPhie\cite{sophie} - GAN \cite{gan} based approach with incorporated social and physical attention mechanisms, Lin - linear pedestrian speed interpolation method that uses the last two observed points, Eq. \ref{lin_u}. %(\ref{lin_V},\ref{lin_u}).

\begin{equation}
    \begin{aligned}
        \Delta{u_i}=u_i^{T_{obs}}-u_i^{T_{obs-1}} \\
        u_i^{t+1} = {u_i^{t}+\Delta{u_i}} 
    \label{lin_u}
    \end{aligned}
\end{equation}
    % delta_step = torch.tensor(indexes_last).reshape(-1,1)
    % speeds = (current_pose - last_visible_pose) / delta_step
\subsection{Results}

In this section, we compare and discuss our proposed method’s performance against mentioned baselines on the ADE $\And$ FDE metrics.
\par
Throughout the following, we report the performance of our approach in multiple configurations.
\textit{Our-nomap}  represents the base model with only two cross-attention blocks used at every encoder block, namely  (i) cross-attention processing embedded agent history positions and (ii) cross-attention processing embedded neighbors positions. 
\textit{Our-resnet} is the model with all three  cross-attention blocks, as shown at figure\ref{fig::encoder}, utilising ResNet-18 as a backbone feature extractor. 
\textit{Our-ViT} is the  model with all three  cross-attention blocks , with ViT\cite{vit} as a backbone feature extractor.
\textit{Ours-patch} is the model with all three cross-attention blocks used at every encoder block, as a backbone we split an image into fixed-size patches and linearly embed each of them, similar to ViT\cite{vit} patching procedure.
\par
Table \ref{table:comparison-eth} shows results of our proposed methods against baselines and current state-of-the-art methods at ETH-UCY datasets. Our proposed method \textit{Ours-ViT} with ViT\cite{vit}  feature extractor achieves $\boldsymbol{18\%}$ boost performance at FDE metric comparing to previous state-of-the-art method. 
Table \ref{table:comparison-sdd} shows results of our proposed method against baselines and current state-of-the-art methods at SDD dataset, \textit{Ours-patch} method achieves superior than previous SOTA ADE results.
% We combined publicly available results of S-LSTM and SoPhie methods, and mesasured performance of PecNet with publicly-available code with deterministic setup (k=1). 
%\textbf{waiting for our metrics}

\begin{table*}[h!]
\caption{Comparison of ADE and FDE results of our method against previously published methods on the ETH \cite{eth} and UCY\cite{ucy} datasets. Both ADE and FDE are reported in  meters.}

% TODO: check datasets
\label{table:comparison-eth}
%\hskip-1.2cm
    % \begin{center}
        \begin{tabular}{|| c | c c c c c | c c c c||} 
         \hline
             Dataset & LSTM\cite{lstm} &
             S-LSTM\cite{social-lstm} &
             S-ATTN\cite{sattn} & Trajectron++\cite{trajectronplusplus} &
             SoPhie\cite{sophie}
             & Ours-nomap & Ours-ResNet & Ours-ViT & Ours-patch\\
             \hline\hline 
             
             ETH\cite{eth} & 1.09/2.41  & 1.09/2.35 & \textbf{0.39}/3.74 & 0.71/1.68 &
             0.70/1.43 & 0.62/\textbf{1.13} & 0.71/1.35 & 0.80/1.20 & 0.64/1.22 \\  
             
             Hotel\cite{eth} & 0.86/1.91 & 0.79/1.76 & 0.29/2.64 & \textbf{0.22}/0.46 &
             0.76/1.67 & 0.29/0.42 & 0.40/0.57 & 0.26/\textbf{0.37} & 0.35/0.65 \\
             
             Univ\cite{ucy} & 0.61/1.31 & 0.67/1.40 & \textbf{0.33}/3.92 & 0.41/1.07 &
             0.54/1.24 & 0.62/0.96 & 0.62/1.03 & 0.66/\textbf{0.92} & 0.70/1.12\\  
             
             Zara 1\cite{ucy} & 0.41/0.88 & 0.47/1.00 & \textbf{0.20}/\textbf{0.52} & 
             0.30/0.77 & 0.30/0.63 &  0.61/1.24 & 0.61/0.70 & 0.53/0.73 & 0.53/1.13 \\
             
             Zara 2\cite{ucy} & 0.52/1.11 & 0.56/1.17 & 0.30/2.13 & \textbf{0.23}/0.59 &
             0.38/0.78 & 0.41/0.70 & 0.44/0.58 & 0.39/\textbf{0.56} & 0.40/0.82\\  
             
             \hline\hline
             Mean: & 0.70/1.52 & 0.72/1.54 & \textbf{0.30}/2.59 & 0.37/0.95 &
             0.54/1.15 & 0.50/0,90 & 0.55/0.84 &  0.52/\textbf{0.75} & 0.53/0.98\\
             \hline
        \end{tabular}
    % \end{center}
\end{table*}

\begin{table*}[h!]
\caption{Comparison of  ADE and FDE results of our method against previously published methods on the SDD\cite{sdd} dataset. Both ADE and FDE are reported in pixels. *measured with deterministic setup (k=1)}

% TODO: check datasets
\label{table:comparison-sdd}
\begin{center}
        \begin{tabular}{|| c | c c c c c |c c c||} 
             \hline
           & PecNet\cite{pecnet} & S-LSTM\cite{social-lstm} &  Lin     & SoPhie\cite{sophie} & DESIRE   & Ours-ResNet & Ours-ViT & Ours-patch \\
             \hline\hline 
      ADE: & 37.70*              & 31.19                    & 19.70    &  16.27             & 19.25    &  15.97 & 15.57   &  \textbf{15.54}\\
             \hline
      FDE: & 88.17*              & 56.97                    & 39.60    & \textbf{29.38} & 34.05        & 31.35  & 30.35   & 30.92\\
             \hline
        \end{tabular}
    \end{center}
\end{table*}

\begin{figure}[ht!]
    \centering
    \includegraphics[width=8.5cm]{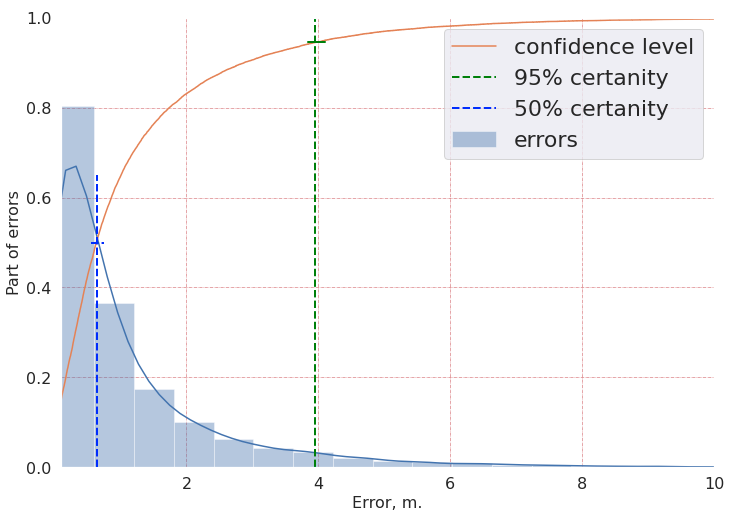}
    \includegraphics[width=8.5cm]{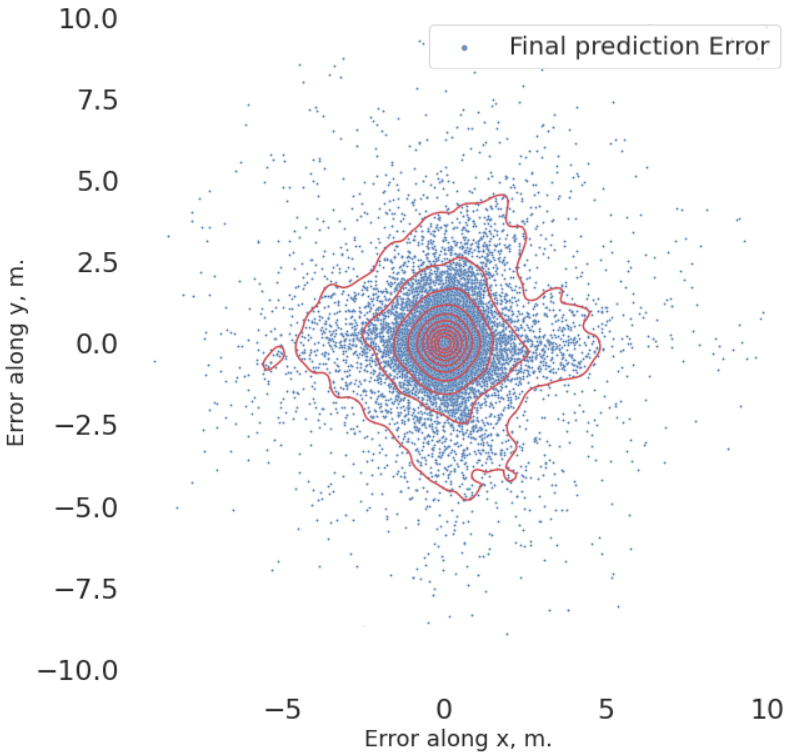}
    \caption{Distributions of final displacement errors on SDD\cite{sdd} dataset. Upper: one-dimensional distribution. Lower: two-dimensional distribution.}
    \label{fig::fde_uncertainty}
\end{figure}

We show the final displacement error distribution and confidence intervals of our method on Figure \ref{fig::fde_uncertainty} for the test data part. Half of all final errors are just in 0.91 meters interval, but we also may observe the long tail of errors resulting from unpredictable human nature.
This final error for the 4.8s prediction interval can be too big for some cases, such as a narrow pedestrian road. That error is smaller for shorter prediction intervals, and the robot will adjust the prediction as it approaches a potential intersection with the pedestrian path.
%for showing that mean of distribution close to zero. \\
% Since we are normalizing the position and orientation of the pedestrian to be predicted at the last observed timestamp in the direction of his movement, the right and the left protrusion of confidence counter throw x ax shows velocity prediction part error. On the other hand, the up and down protrusion of confidence counter throw y ax show moving direction miss prediction part.

% \begin{figure}[h]
%     \centering
%     \includegraphics[width=6.1cm]{figures/fde_2d_distrib_new2.png}
%     \caption{CICAT 2d distribution of SDD\cite{sdd} final displacement errors. \gf{explain how this graphic was obtained and what it means}}
%     \label{fig::fde_2d_distrib.png}
% \end{figure}

We visualize two-dimensional final error distribution with confidence iso-contours on Figure \ref{fig::fde_uncertainty}, where each point of that plot is an error, in normalized (according to explained in section III) coordinate system, between ground truth and predicted agent final position from SDD test data part.
There are two types of errors that can be distinguished in Figure \ref{fig::fde_uncertainty}.
The first type of error is contour protrusions along the X-axis.
The second type of error is contour protrusions on the upper and lower parts of the error distribution.
Since we are normalizing the position and orientation of the pedestrian to be predicted at the last observed timestamp in the direction of his movement, the first type of error shows part of the errors when the model correctly predicted the direction of motion but did not correctly predict the value of the person's movement speed. 
The second type of error shows scenarios when the model did not correctly predict the direction of movement, which may be associated with unexpected changes in the person's direction of movement. This type of behavior can be predicted by producing a set of possible trajectories of a person's movement instead of a single most likely trajectory.

\subsection{Implementation details}
All the sub-networks used in proposed model are Multi-Layered Perceptrons with ReLU non-linearity. 
Latent array is a learnable array of shape 12x48,
$MLP_{pose}$ is a stacked linear layers with architecture: 2 $\rightarrow$ 8 $\rightarrow$ 32. 
$MLP_{goal}$ is a stacked linear layers with architecture: 576 (48*12) $\rightarrow$ 256 $\rightarrow$ 64 $\rightarrow$ 2.
$MLP_{traj}$ is a stacked linear layers with architecture: 50 $\rightarrow$ 256 $\rightarrow$ 64 $\rightarrow$ 24.
Parameters of Multihead Attention layer is next: number of parallel attention heads is 8, embedding dimension is 48 (32 for every pose + 16 of position embedding for every of 12 known positions). The number of sequential encoder blocks used in our model is 4.
The entire network is trained end to end with $\mathcal{L}_{model} = \mathcal{L}_{ADE} + \lambda \cdot \mathcal{L}_{FDE}$ loss with $\lambda$ = 0.5. Parameters optimization was performed using an ADAM optimizer with a batch size of 32 and initial learning rate of $5e^{-4}$. We decay the learning rate by factor of 0.2  after every 30 epochs. Total number of trained epochs is 65.

\section{Conclusion}

In this paper, we have proposed a trajectory prediction algorithm conditioned by multiple sources of input data.
Our approach first creates the embeddings of data from map images using a pre-trained backbone network and the past history of agents from a multilayer perceptron.
Our main contribution is on how to combine the high dimensional embeddings from data into an approach that uses cross-attention and transformers iteratively, allowing to effectively capture the complex relations between the ego-agent prediction and the environment.

Our proposed method results in simplified network architecture, more flexible for further configurations or modifications while its performance is similar to other SoTA approaches, being best at some sequences.

\end{document}